\documentclass[11pt,a4paper]{article}
\usepackage[hyperref]{acl2021}
\usepackage{times}
\usepackage{latexsym}
\usepackage{amsmath,amssymb,amsopn,amsthm,bm,bbm}
\usepackage{booktabs}
\usepackage{graphicx}

\usepackage{float}
\usepackage{todonotes}

\usepackage{microtype}

\aclfinalcopy 
\def\aclpaperid{11} 

\title{Error-Aware Interactive Semantic Parsing of OpenStreetMap}

\author{Michael Staniek \\
     Computational Linguistics\\
  Heidelberg University\\
  Germany\\
   \small\texttt{staniek@cl.uni-heidelberg.de}\\\And
  Stefan Riezler \\
   Computational Linguistics \& IWR\\
  Heidelberg University\\
  Germany\\
   \small\texttt{riezler@cl.uni-heidelberg.de}}
\date{}

\begin{document}
\maketitle
\begin{abstract}
In semantic parsing of geographical queries against real-world databases such as OpenStreetMap (OSM), unique correct answers do not necessarily exist. Instead, the truth might be lying in the eye of the user, who needs to enter an interactive setup where ambiguities can be resolved and parsing mistakes can be corrected.
Our work presents an approach to interactive semantic parsing where an explicit error detection is performed, and a clarification  question is generated that  pinpoints  the  suspected  source  of  ambiguity  or error and communicates it to the human user. Our experimental results show that a combination of entropy-based uncertainty detection and beam search, together with multi-source training on clarification question, initial parse, and user answer, results in improvements of 1.2\% F1 score on a parser that already performs at 90.26\% on the NLMaps dataset for OSM semantic parsing.
\end{abstract}

\section{Introduction}

Semantic Parsing has the goal of mapping natural language questions into formal representations that can be executed against a database. 
If real-world large-scale databases such as OpenStreetMap (OSM)\footnote{\url{www.openstreetmap.org}} need to be accessed, the creation of gold standard parses by humans can be complicated and requires expert knowledge, and even reinforcement learning from answers might be impossible since unique correct answers to OSM queries do not necessarily exist. Instead, uncertainties can arise due to open-ended lists (e.g., of restaurants), fuzzily defined geo-positional objects (e.g., objects ``near'' or ``in walking distance'' of other objects), or by ambiguous mappings of natural language to OSM tags\footnote{For example, \emph{recreation grounds} can map to tags reserved for \emph{leisure} purposes or for official \emph{landuse} registration; \emph{bars} map to tags \emph{bar} and \emph{pub} that differ in that only the latter sells food; \emph{off-license} shops can have licenses to sell only \emph{wine} or all kinds of \emph{alcohol}.}, with the truth lying in the eye of the beholder who asked the original question. Semantic parsing against OSM thus asks for an interactive setup where an end-user inter-operates with a semantic parsing system in order to negotiate a correct answer, or to resolve parsing ambiguities and to correct parsing mistakes, in a dialogical process.  

Previous work on interactive semantic parsing \citep{LabutovETAL:18,YaoETAL:19,ElgoharyETAL:20} has put forward the following dialogue structure: i) the user poses a natural language question to the system, ii) the system parses the user question and explains or visualizes the parse to the user, iii) the user generates natural language feedback, iv) the parser tries to utilize the user feedback to improve the parse of the original user question. In most cases, the ``explanation" produced by the system is restricted to a rule-based reformulation of the parse in a human intelligible form, whereas the human user has to take guesses about where the parse went wrong or is ambiguous.

The goal of our paper is to add an explicit step of error detection on the parser side, resulting in an automatically produced clarification question that pinpoints the suspected source of ambiguity or error and communicates it to the human user.
Our experimental results show that a combination of entropy-based uncertainty detection and beam search for differences to the top parse yield concise clarification questions. We create a dataset of 15k clarification questions that are answered by extracting information from gold standard parses, and complement this with a dataset of 960 examples where human users answer the automatically generated questions. Supervised training of a multi-source neural network that adds clarification questions, initial parses, and user answers to the input results in improvements of 1.2\% F1 score on a parser that already performs at 90.26\% on the NLMaps dataset for OSM semantic parsing.

\section{Related Work}

\citet{YaoETAL:19} interpret interactive semantic parsing as a slot filling task, and present a hierarchical reinforcement learning model to learn which slots to fill in which order. They claim the automatic production of clarification questions by the agent as a main feature of their approach, however, what is actually used in their work is a set of 4 predefined templates.
\citet{ElgoharyETAL:20} show an interpretation of the parse that is understandable for laypeople with a template-based approach, and present different approaches to utilize the user response to improve the parser. In their work, the explantion on the parser side is purely template-based, whereas our work explicitly informs the clarification question by possible sources of parse ambiguities or errors.

Considerable effort has been invested in the creation of large datasets for parsing into SQL representations.
\citet{yu-etal-2018-spider} created a dataset called Spider which is a complex, cross-domain semantic parsing and text-to-SQL dataset. Their annotation process was very extensive, and involved 11 computer science students who invested a total of 1,000 hours into asking natural language queries and creating the corresponding SQL query. Extensions of the Spider dataset, SParC \citep{sparc}, or Co-SQL \citep{ yu2019cosql}  involved even more computer science students. Our work attempts an automatic construction of concise clarification questions, allowing for faster dataset construction.

\section{(Multi-Source) Neural Machine Translation}

Our work employs as a semantic parser a sequence-to-sequence neural network \citep{seq2seq} that is based on an recurrent encoder and decoder architecture with attention \citep{BahdanauETAL:15}.

Given a corpus of aligned data $\mathrm{D}=\{(x_n, y_n)\}_{n=1}^N$ of user queries $x$ and semantic parses $y$, standard supervised training is performed by minimizing a Cross-Entropy objective
$-\frac{1}{N}\sum_{n=1}^N\sum_{t=1}^T \log p(y_{n,t}|y_{n,<t}, x_n)$,
where the probability of the full output sequence $y = y_1, y_2, ..., y_n$ is calculated by the product of the probability for every timestep where $p(y|x) = \prod_{t=1}^T p(y_t|y_{<t}, x)$. 

This model can be easily extended to multi-source learning \citep{ZophKnight:16} by using not only one, but multiple encoders. This means that there are actually multiple sequences of hidden states. The decoder hidden state is consequently initialized by a linear projection of the average of the last hidden states of all encoders $c=\frac{1}{N}\sum_{i=1}^Nh_iW_l$, and needs to implement a separate attention mechanism for every encoder.

To be able to fine-tune a model with feedback from a user, the standard cross-entropy objective cannot be used because the desired target is not a gold parse, but a parse $\Tilde{y}$ predicted by the system, that has been annotated with positive and negative markings by a human user. This can be formalized as assigning a  reward $\delta_t$ that is either positive or negative to every token in the parse ($\delta_{t+}=0.5$ and $\delta_{t-}=-0.5$). It is then possible to maximize the likelihood of the correct parts of the parse by optimizing a weighted supervised learning objective $\underset{x,\Tilde{y}}{\sum} \underset{t=1}{\overset{T}{\sum}}\delta_t\log p(\Tilde{y_t}|x, y_{<t})$. \cite{petrushkov-etal-2018-learning}

\section{Neural Semantic Parsing of OSM}

\subsection{Data}

Our work is based on the NLmaps v2 dataset.\footnote{\url{www.cl.uni-heidelberg.de/statnlpgroup/nlmaps/}}
NLmaps builds on the Overpass API which allows the querying of the OSM database with natural language queries. 
This dataset includes template-based expansions leading to duplicates in train and test sets. 
However, these expansions introduced problematic features into the data in that OSM tags were inserted which, according to the documentation in the OSM developer wiki, should not be used:

\begin{itemize}
    \item Is there Recreation Grounds in Marseille
    \setlength{\itemsep}{-5pt}
    \item[$\rightarrow$] \small{\texttt{query(area(keyval('name','Marseille')),\\nwr(keyval('\textbf{leisure}','recreation\_ground'),\\qtype(least(topx(1))))}}
    \setlength{\itemsep}{2pt}
    \item Recreation Ground in Frankfurt am Main
    \setlength{\itemsep}{-5pt}
    \item[$\rightarrow$]
    \small{\texttt{query(area(keyval('name','Frankfurt am Main')),\\nwr(keyval('\textbf{landuse}','recreation\_ground')),\\qtype(latlong))}}
\end{itemize}

While \textit{leisure=recreation\_ground} certainly exists as a tag\footnote{\url{https://wiki.openstreetmap.org/wiki/Tag:leisure\%3Drecreation\_ground}.}, its use is heavily discouraged\footnote{{\url{https://wiki.openstreetmap.org/wiki/Tag:landuse\%3Drecreation\_ground}}.}. Furthermore, several mistakes were introduced in the data by the augmentation with the help of a wordlist. For example, an automatically generated natural language question based on this wordlist asks for bars, whereas the gold parse associated to that question asks for pubs instead:

\begin{itemize}
    \item Where Bars in Bradford
    \setlength{\itemsep}{-5pt}
    \item[$\rightarrow$] \small{\texttt{query(area(keyval('name','Bradford')),\\nwr(keyval('amenity',\textbf{'pub'})),\\qtype(latlong))}}\
\end{itemize}

Conceptually, bars and pubs may not be that different to each other, but OSM advises a strict distinction between bars and pubs \footnote{{\url{https://wiki.openstreetmap.org/wiki/Tag:amenity\%3Dbar}}.}. While a pub sells alcohol on premise, a pub also sells food, the athmosphere is more relaxed and the music is quieter compared to a bar.

Lastly, ambiguity was introduced because natural language words now map to multiple different OSM tags. This leads to the following data occurrences:

\begin{itemize}
    \item shop Off Licenses in Birmingham 
    \setlength{\itemsep}{-5pt}
    \item[$\rightarrow$] \small{\texttt{query(area(keyval('name','Birmingham')),\\nwr(keyval('shop','\textbf{alcohol}')),
    \\qtype(findkey('shop')))}}
    \setlength{\itemsep}{2pt}
    \item How many closest Off License from Wall Street in Glasgow
    \setlength{\itemsep}{-5pt}
    \item[$\rightarrow$] \small{\texttt{query(around(center(area(\\keyval('name','Glasgow')),\\nwr(keyval('name','Wall Street'))),\\search(nwr(keyval('shop','\textbf{wine}'))),\\maxdist(DIST\_\\INTOWN),\\topx(1)),qtype(count))}}
\end{itemize}

The previous examples show that for the same keyword "Off License" both \textit{shop=alcohol} and \textit{shop=wine} are valid interpretations.

Finally, since the data was augmented first, and only afterwards split into train, development and test sets, there is a lot of overlap between the train and test data. This is problematic because a proper evaluation should also test for overfitting, which does not work if data is shared between different splits, as shown in the following examples:

\begin{itemize}
    \item Train: cinema in Nantes
    \item Dev: cinema in Paris
    \item Test: cinemas in Paris
\end{itemize}

We applied a dataset de-duplication by removing all datapoints from the development and test sets which are identical to training datapoints when location (e.g., \textit{Paris}) and POI (e.g., \textit{cinema}) are masked. This results in the dataset described in table \ref{tab:dia_set}.

\subsection{Semantic Parsing}

We use the Joey NMT \citep{kreutzer2019minimalist} as framework to build a baseline parser. The basic Joey NMT architecture is modified to allow for a multi-source setup (see Figure \ref{fig:multisource} in the appendix) and for learning from markings.\footnote{Meta-parameter settings are reported in the appendix.} 

As evaluation metrics we use exact match accuracy, defined as $\frac{1}{N}\sum_{n=1}^N\delta(\text{predicted, gold})$  of a predicted parse and the gold parse. Furthermore, we report F1 score as harmonic mean of recall, defined as the percentage of fully correct answers divided by the set size, and precision, defined as the percentage of correct answers out of the set of answers with non-empty strings.
 
A character-based Joey NMT semantic parser is able to improve the results reported in \citet{LawrenceRiezler:18} on the dataset without de-duplication, as shown in Table \ref{tab:old_res}. All results presented in the following are relative improvements over our own baseline parser, reported on the de-duplicated dataset for which no external baseline is available.

  \begin{table}[t!]
     \centering
     \begin{tabular}{ll}
     \toprule
          System & F1 \\
          \midrule
          \citet{lawrencethesis} & 80.36 \\
          \citet{lawrencethesis}+NER & 90.09 \\
          token-based & 83.43 \\
          character-based & 93.77 \\
          \bottomrule
     \end{tabular}
     \caption{F1 results of single-source models on the original NLmaps v2 dataset.}
     \label{tab:old_res}
 \end{table}
 
\section{Generation of Clarification Questions}

On of the goals of error-aware interactive semantic parsing is to alert to user about suspected sources of ambiguity and error by initiating a dialogue. The parser thus needs to detect uncertainty in its output, and generate a clarification questions on the detected source of uncertainty. We use entropy-based uncertainty measures. Firstly, entropy per timestep $t$ is measured as $-\sum_{\Tilde{y_t}}p(\Tilde{y_t}|x, y_{<t}) \log p(\Tilde{y_t}|x, y_{<t})$. This is employed to calculate the entropy of a token as the mean of the character entropies for each of a token's characters.\footnote{A visualization of entropy is reported in the appendix.}
Based on entropy information, we generate simple questions by employing a template-based method which incorporates the least certain token: "Did you mean \$token?". 
Furthermore, we offer alternative answers for the user based on beam search of size 2. 
This heuristic is justified experimentally since always taking the first beam yields an accuracy of 92.7\%, while another 5\% of accuracy can be gained by choosing the second beam. This verifies the usefulness of proposing entries in the second beam as alternative in clarification questions: "Did you mean \$token or \$alternative?". 

\begin{figure}[t!]
    \centering
    \includegraphics[width=0.5\textwidth,height=0.405\textheight]{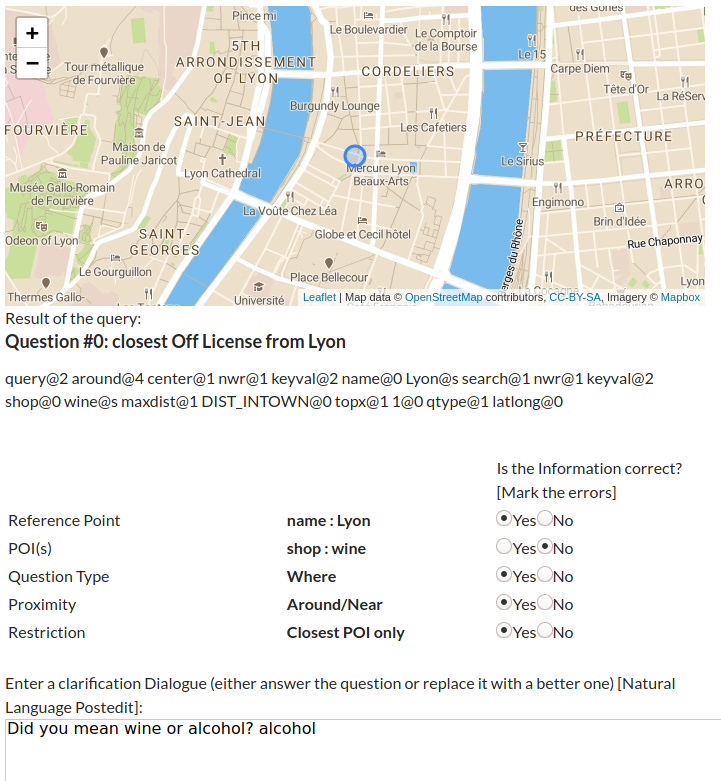}
    \caption{Annotation setup for human interaction study.}
    \label{fig:annotation}
\end{figure}

  \begin{table}[t!]
     \centering
     \begin{tabular}{ccc}
     \toprule
          System & Accuracy & F1 \\
          \midrule
          baseline & 83.50 &  90.26\\
          baseline + hyps & 83.66 & 90.85 \\
          baseline + dia & 84.74 & 92.02 \\
          baseline + hyps + dia & 84.84 & 91.47 \\
          \midrule
          baseline + hyps + dia + log & 85.01 & 91.61 \\
          \bottomrule
     \end{tabular}
     \caption{Results of the multi-source models compared to the single-source model taking only the source into account on the modified test data.
     }
     \label{tab:test_res}
 \end{table}
 
  \begin{figure*}[t!]
    \centering
    \includegraphics[width=1\textwidth,height=0.5\textheight]{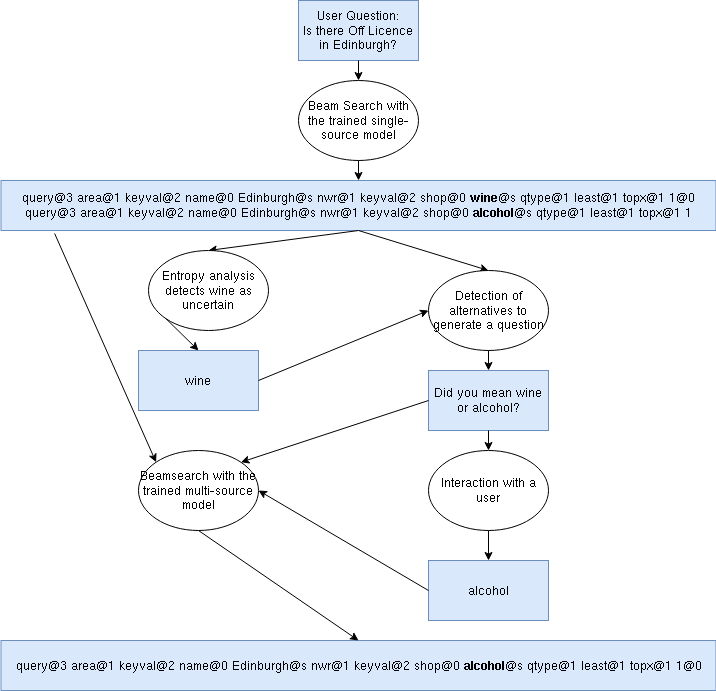}
    \caption{Workflow for the interaction process.}
    \label{fig:annotation2}
\end{figure*}
 
 \section{Experiments on Synthetic Dialogues}

In a first experiment, we generated entire dialogues synthetically, that is, the clarification question from the parser and synthetic user answers. The latter were constructed by checking if either the original token or the alternative is contained in the given gold parse. Dataset statistics for train, development and test splits are given in Table \ref{tab:dia_set}.

\begin{table}[t!]
    \centering
    \begin{tabular}{ll}
    \toprule
         Split & Count \\
         \midrule
         Train & 15,658\\
         Dev &  961\\
         Test  & 4156 \\
         \bottomrule
    \end{tabular}
    \caption{Statistics of dialogue-enriched data.}
    \label{tab:dia_set}
\end{table}

Model training is performed by extending the character-based baseline model by additional encoders for the dialogue (question and answer) and the predicted parse hypothesis.
Experiments show that the character-based multi-source model including hypothesis and dialogue as additional input (line 4)  outperforms the baseline (line 1) by more than 1 point in accuracy and F1 score (Table \ref{tab:test_res}). 
This difference is statistically significant with a p-value of 0.0483 determined by approximate randomization.

\section{Human Interaction Study}

We furthermore performed a small field study where human users interacted with the system. Parses for queries from both train and development parts of the dataset were generated and augmented with automatically created clarification questions based on the uncertainty model. Examples were then filtered to keep only those parses that contained a parse mistake or parse ambiguity. This resulted in a total of 930 annotation tasks.\footnote{Both synthetic and user data will be publicly released.}

The annotation interface shown in Figure \ref{fig:annotation} illustrates the system-user interaction: Human annotators are presented with a natural language query ("closest Off License from Lyon"), the parse (shown below in linearized form), and the result of the generated parse (show as the map extract on top of the figure). In addition to the linearized form of the predicted parse, a human-intelligible list format of the key-value pairs in the parse\footnote{\url{https://wiki.openstreetmap.org/wiki/Map_Features}} is presented, following the annotation interface of \citet{LawrenceRiezler:18}. The task of the human users is to mark the errors in the list of keys and values, and to answer or correct the clarification question. The markings are used as feedback in the weighted fine-tuning objective of \citet{petrushkov-etal-2018-learning}. 
As the outputs of the model are on character-level, the token-level reward of the annotations is distributed onto them for training. The final model is trained on the weighted objective in a multi-source fashion, taking parse hypothesis, clarification question, and logged user answer as additional inputs.
Line 5 in Table \ref{tab:test_res} shows that fine-tuning a multi-source model that takes hypothesis, dialogue, and logged answer as additional input increases the sequence accuracy by another ~0.15\%. 
This difference is statistically significant with a p-value of 0.0027 determined by approximate randomization.
The interaction process can be seen in Figure \ref{fig:annotation2}.\footnote{Additional experiments using the human annotations as test data are reported in the appendix.}

\section{Conclusion}

Ambiguities or errors in real-world semantic OSM parsing arise because of different tagging preferences of developers and users, an issue that can only be solved by an interactive setup where a parser is aware of its errors, and a satisfactory answer is found by the user marking parse errors and communicating alternatives. Our current work is a first step towards precise communication and offline learning in interactive semantic parsing. An interesting future direction of work is to move to online learning in interactive semantic parsing.

\subsubsection*{Acknowledgments}
We would like to thank Christian Buck and Massimiliano Ciaramita for initial fruitful discussions about this work. We would like to thank Christian Buck and Massimiliano Ciaramita for initial fruitful discussions about this work. The research reported in this paper was supported by a Google Focused Research Award on "Learning to Negotiate Answers in Multi-Pass Semantic Parsing". 

\bibliographystyle{acl_natbib}
\bibliography{acl2021}

\newpage
\newpage

\appendix
 
 \begin{table}[b!]
     \centering
     \begin{tabular}{c|c|c|c}
          Parameter & \citet{LawrenceRiezler:18} & token-based & character-based \\
          \hline
          Attention mechanism & bahdanau & bahdanau & bahdanau\\ 
          RNN type & gru & gru & gru\\ 
          Embedding size & 1000 & 620 & 620 \\
          Encoder layer count & 1 & 1 & 1\\ 
          Encoder hidden size & 1024 & 400 & 400 \\ 
          Decoder layer count & 1 & 1 & 1 \\ 
          Decoder hidden size & 1024 & 800 & 800 \\ 
     \end{tabular}
     \caption{Parameter overview compared to \citet{LawrenceRiezler:18}.}
     \label{tab:hyperparameter}
 \end{table}

  \begin{figure}[b!]
    \centering
    \includegraphics[width=0.8\textwidth,keepaspectratio]{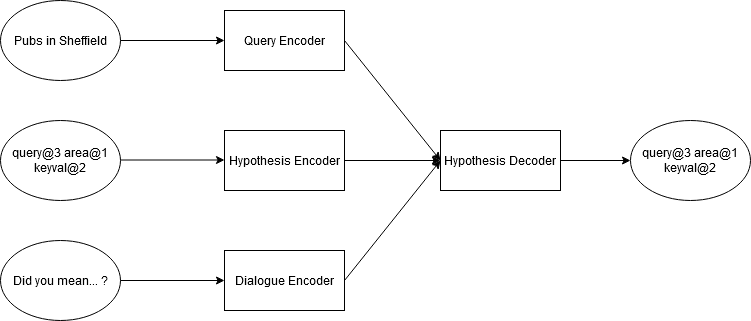}
    \caption{Multi-source semantic parsing.}
    \label{fig:multisource}
\end{figure}

\begin{figure}[b!]
    \centering
    \includegraphics[width=1.4\linewidth,keepaspectratio]{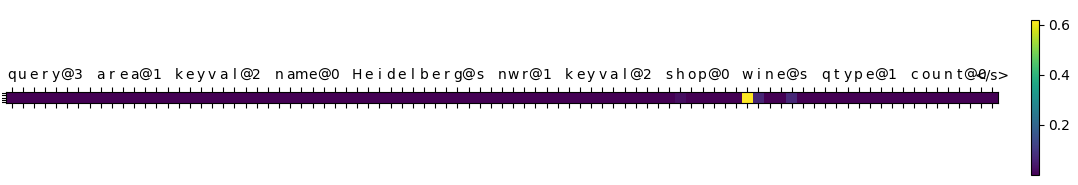}
    \caption{Character entropy for parse of query "How many Off License in Heidelberg".}
    \label{fig:entropy_char}
\end{figure}

\section{Supplementary Material for "Towards Error-Aware Interactive Semantic Parsing"}
\subsection{Hyperparameter Settings}

 \subsection{Evaluation on the human annotated data}
 
   \begin{table}[h!]
     \centering
     \begin{tabular}{c|c|c}
          System & Accuracy & F1 \\
          \hline
          char & 0 & 43.07 \\
          char+ hyps + dia & 25.09 & 60.85
     \end{tabular}
     \caption{Test results on human-annotated data.}
     \label{tab:on_log_test}
 \end{table}
 
 In an additional experiment, we evaluated the models that were trained on the synthetically generated dataset on the data resulting from the human interaction study. The result of comparing the baseline model with the multi-source model trained on parse hypothesis and synthetic dialogue as additional inputs is shown in Table \ref{tab:on_log_test}. The astonishing gains of over 15\% in F1 score can be explained by the fact that the data for human annotation set were filtered to include only examples for which the baseline parser did not match the gold standard parse (thus producing an accuracy score of $0$).

\subsection{Entropy visualization}

The entropy of the parse of the sentence "How many Off License in Heidelberg" can be seen in Figure \ref{fig:entropy_char}. The character-based model shows uncertainty with respect to the token \textit{wine}. This is the desired result because the alternative for this position would be \textit{alcohol}.

\end{document}